\documentclass[conference]{IEEEtran}
\IEEEoverridecommandlockouts
\usepackage{cite}
\usepackage{amsmath,amssymb,amsfonts}
\usepackage{algorithmic}
\usepackage{graphicx}
\usepackage{textcomp}
\usepackage{xcolor}
\def\BibTeX{{\rm B\kern-.05em{\sc i\kern-.025em b}\kern-.08em
    T\kern-.1667em\lower.7ex\hbox{E}\kern-.125emX}}
\usepackage{booktabs, floatrow, makecell}
\makeatletter
\let\NAT@parse\undefined
\makeatother
\usepackage{tikz}
\newcommand\copyrighttext{%
  \footnotesize \textcopyright 2021 IEEE.  Personal use of this material is permitted.  Permission from IEEE must be obtained for all other uses, in any current or future media, including reprinting/republishing this material for advertising or promotional purposes, creating new collective works, for resale or redistribution to servers or lists, or reuse of any copyrighted component of this work in other works.}
\newcommand\copyrightnotice{%
\begin{tikzpicture}[remember picture,overlay]
\node[anchor=south,yshift=10pt] at (current page.south) {\fbox{\parbox{\dimexpr\textwidth-\fboxsep-\fboxrule\relax}{\copyrighttext}}};
\end{tikzpicture}%
}
\usepackage[colorlinks=true,urlcolor=blue,linkcolor=green,backref=page]{hyperref}
\begin{document}

\title{A Neural Anthropometer Learning from Body Dimensions Computed on Human 3D Meshes}

\author{\IEEEauthorblockN{Yansel González Tejeda}
\IEEEauthorblockA{\textit{Computer Science Department} \\
\textit{Paris Lodron University of Salzburg}\\
Austria \\
https://orcid.org/0000-0003-1002-3815}
\and
\IEEEauthorblockN{Helmut A. Mayer}
\IEEEauthorblockA{\textit{Computer Science Department} \\
	\textit{Paris Lodron University of Salzburg}\\
	Austria \\https://orcid.org/0000-0002-2428-0962}
}

\maketitle
\copyrightnotice

\begin{abstract}
Human shape estimation has become increasingly important both theoretically 
and practically, for instance, in 3D mesh estimation, distance garment 
production and computational forensics, to mention just a few examples. As 
a further specialization, \emph{Human Body Dimensions Estimation} (HBDE) 
focuses on estimating human body measurements like shoulder width or chest 
circumference from images or 3D meshes usually using supervised learning 
approaches. The main obstacle in this context is the data scarcity problem, 
as collecting this ground truth requires expensive and difficult 
procedures.
This obstacle can be overcome by obtaining realistic human measurements 
from 3D human meshes. However, a) there are no well established methods to 
calculate HBDs from 3D meshes and b) there are no benchmarks to fairly 
compare results on the HBDE task. Our contribution is twofold. On the one 
hand, we present a method to calculate 
right and left arm length,
shoulder width, and inseam (crotch height) from 3D meshes with focus on 
potential medical, virtual try-on and distance tailoring applications.
On the other hand, we use four additional body dimensions calculated 
using recently 
published 
methods to assemble a set of eight body dimensions which we use as a 
supervision 
signal to our Neural Anthropometer: a convolutional neural network capable 
of estimating these dimensions. To assess the estimation, we train the 
Neural Anthropometer with synthetic images of 3D meshes, from which we 
calculated the HBDs and 
observed that the network's overall mean estimate error is $20.89$ mm 
(relative error of 2.84\%). The results we present are fully 
reproducible and establish a fair baseline for research on the task of 
HBDE, therefore enabling the community with a valuable method.
\end{abstract}


\section{Introduction}

\begin{figure*}
	\begin{center}
		\includegraphics{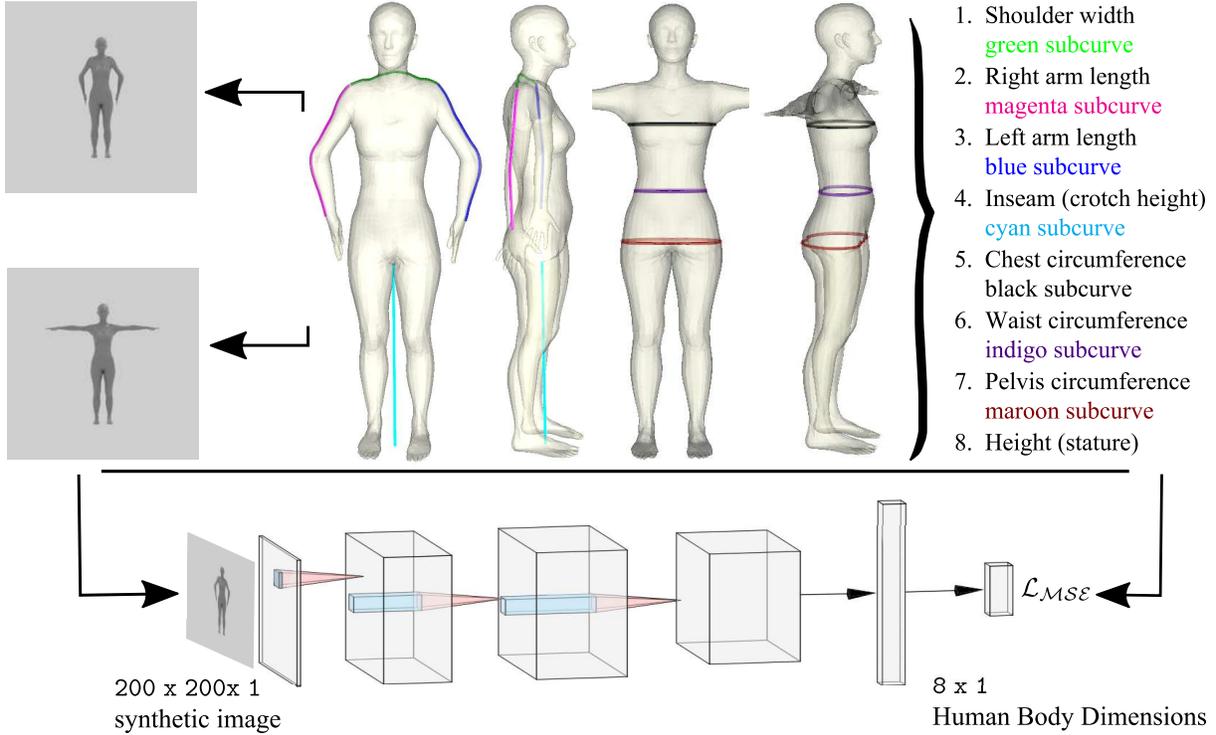}
	\end{center}
	\caption{The Neural Anthropometer framework: a 
		CNN that is able to estimate human body dimensions from synthetic 
		images of persons in two poses. We propose to address the 
		problem of data scarcity in estimating human body dimensions by 
		establishing methods to compute them from 3D meshes. Top center: we 
		present a method to calculate shoulder width, right and left arm length 
		and inseam (crotch height). In addition to these human dimensions, we 
		use Calvis \cite{ygtham2020calvis} to augment the set with four more 
		measurements plus height (top right). We then synthesize grayscale 
		200 pixels square images ($\mathtt{200 \, x \, 200 \, x \,1}$) from 3D 
		female and male (not shown) meshes in two poses to use them as 
		input to a CNN (top left). Bottom: The Neural Anthropometer can be fed 
		with these synthetic images. The 
		supervision signal is a vector of eight human 
		body 
		dimensions and the loss is the Mean Square Error $\mathcal{L_{MSE}}$ 
		between the actual and 
		the estimated measurements. The network architecture is described in 
		subsection \ref{subsec:na_arqui}.}
	\label{fig:na_overview}
\end{figure*}
Human shape analysis has been, historically, an active area of research. 
Recently, as a further specialization, \emph{human shape estimation} (HSE) has 
been established as a solid scientific field whose main concern is to 
investigate methods to infer the 
shape of a human body from underconstrained information.

Depending on these methods' input, HSE scenarios can be categorized as 
estimation from imagery (2D) or point clouds or meshes (3D). Regarding the 
output, there are two tasks commonly performed, namely a) 
parameter estimation, e.g., estimation of SMPL 
(Skinned Multi-Person Linear) model parameters; and b) estimation of 
a surface that best explain the input image, a.k.a. 3D 
reconstruction, or more recently, human mesh recovery.

However, the task of human body dimensions estimation (HBDE)
has, to date, received significantly less attention. Human body dimension (HBD)
refers in this context to anthropometrical measurements, e.g., shoulder width 
or crotch height. Predicting these 3D human body measurements from images is 
fundamental in several scenarios like virtual try-on, animating, 
ergonomics, computational forensics and even health and mortality estimation 
\cite{Heymsfield2018}.

The HBDE problem given a
single image is an inverse (underconstrained) problem. If we consider the 
human body dimensions 
as the `cause' and the images of these humans as the observations, then our 
goal is to calculate from a set of observations the causal factors. 
Furthermore, information gets lost
when a camera is used to capture the human body in 3D space to form a 2D
image.
This problem may be tackled by a supervised learning approach, 
specifically, convolutional neural networks (CNNs). The fundamental challenge 
of this approach is to overcome the data scarcity 
problem. A moment of thought 
reveals the extraordinary effort demanded by the collection of these data: 
thousands of people would have to be measured and photographed. While such a 
survey has been conducted in the 
past (CAESAR project \cite{robinette1999caesar}), we need to investigate 
different approaches that are not confronted with this heavy burden. 
Accordingly, we explore an alternative method 
where HBDs are not derived from real people, but from models of real humans.

Generating 3D realistic human bodies and performing anthropometrical 
measurements 
on 
them is by no means a trivial task. Despite practitioners having a general 
idea, there is no consens regarding how each of these measurements (if any) 
should be 
measured, neither in 3D models nor even in humans. Moreover, in the rare 
case that the procedure to measure a particular HBD is well established among 
anthropometrists and tailors, it often refers to 
human body landmarks (HBL) and anatomical joints (AJ). For example, the 
European clothing standard EN 13402 defines arm length as ``distance, measured 
using the tape measure, from the armscye/shoulder line intersection (acromion), 
over the elbow, to the far end of the prominent wrist bone (ulna), with the 
subject's right fist clenched and placed on the hip, and with the arm bent at 
90°''. These points of reference (HBL and AJ) pose a challenging scenario for 
computer vision systems. On the one side HBL are not well defined and on the 
other side AJ are not visible.

Yet another problem is the comparability of results related to HBDE. Since the 
methods to compute HBDs are not consistent, the estimation results can not 
be fairly compared. For example, \cite{kilic2016evaluating} examined four 
garment pattern preparing systems and found that HBDs differed entirely among 
them and all violated anthropometric rules. We are keen to advance the field in 
this matter by making our method's implementation publicly available.

Let us turn now to the problem of HBDE.

\subsection{The Problem of Estimating Human Body Dimensions}

Our goal is, given an image $\mathcal{I}$ from a 3D 
human body with HBD $D$, to return a set of estimated human body 
dimensions. Let us denote the 
estimated dimensions vector as $\hat{D}$, then our problem is to train 
a CNN $\mathcal{M}$ that has a good generalization performance such that (at 
least 
approximately)

\begin{equation}  
	\hat{D} = \mathcal{M}(\mathcal{I}(D))
\end{equation}

Elements of the human dimensions vector $D$ are, in general, body measurements 
like 
height, shoulder width, and waist circumference, or other shape information 
like 
body 
surface area, weight, or body fat percentage, to mention just a few examples. 
The elements of $\hat{D}$ are their estimated values. In this paper, we 
focus on eight of these human body dimensions, namely, shoulder width, right 
and 
left 
arm length and inseam (crotch height); chest, waist and pelvis circumference 
and height (stature).

We assume that we draw the dataset from the generating distribution 
$\mathcal{G}$. The model $\mathcal{M}$ minimizes the probability, 
$C(\mathcal{M})$, of making a 
wrong 
prediction:

\begin{equation}  
	C(\mathcal{M}) = \Pr_{(\mathcal{I}, D) \sim \mathcal{G}} 
	[\mathcal{M}(\mathcal{I}) \neq D]
\end{equation}

Note that this is a deep regression (in contrast to a classification) multitask 
learning problem \cite{zhen.2018}. The 
term 
deep regression has become popular over the past few years in the context of 
deep learning. However, the lack of its 
systematic evaluation has been recently criticized 
\cite{DBLP:journals/corr/abs-1803-08450}.

\subsection{Contributions and scope}
Our key contribution is the Neural Anthropometer approach: a method to obtain 
eight HBDs from 3D human meshes that we use as a supervision signal to conduct 
HBDE with a CNN whose inputs are synthetic images from those meshes (Fig. 
\ref{fig:na_overview}).

We believe that our fully reproducible method to obtain 
ground truth provides the scientific community with a valuable methodology for 
HSE 
in general and particularly for HBDE. Moreover, the Neural Anthropometer 
estimation results establish a solid baseline that researchers can use 
to compare future work to. Upon publication, we will make our code publicly available for research purposes under \url{https://github.com/neoglez/neural-anthropometer}.

In this paper, we will first review briefly the state of art, as well as our 
approach to compute human body dimensions from 3D models and use them for 
supervised learning. We will then present one experiment that illustrates the 
plausibility of our approach. Finally, we will discuss important aspects of 
our findings.

\section{Related Work}

In this section we present research relevant to our goal. We group related work 
around topics of interest in HSE.

\subsection{Human Body Dimensions Estimation from Images.}\label{subsec:hbdefi}

Previous research has used HBDs directly recovered from individuals to estimate 
body measurements from images \cite{ams.2008.BenAbdelkaderY08, Sigal.2008}. 
Height, arm span and even weight have been employed as ground truth. Chest and 
waist size have been considered as well \cite{Guan.2013}. In contrast, we do 
not conduct measurements directly on humans, but calculate them  
consistently from 3D meshes.
Other researches focus on an small set of dimensions or even only one. For 
example, \cite{Sriharsha2019} estimates height in surveillance areas. A set of 
three HBDs (chest, waist and pelvis circumference) is defined by 
\cite{ygtham2020calvis}. We augment this set 
with four more dimensions.

Another significant research direction studies shape as parameters of some 
model. For example \cite{Bogo:ECCV:2016} claim 
describing the first method 
to 
estimate pose and shape from a single image, while \cite{DBLP:Lassner2017} 
estimates shape from 91 keypoints using decision 
forests. Shape is understood as SMPL model \cite{Loper.2015} parameters but not 
human dimensions. 
In general, previous work has concentrated in parameter or 3D mesh estimation 
(mesh recovery). In contrast, we focus in estimating human body 
measurements in an end-to-end manner.

\subsection{Human Body Data Synthesis.}\label{subsec:hbds}

The Shape Completion and Animation of People (SCAPE) model\cite{Anguelov.2005} 
opened wide possibilities in the field of 
human shape estimation. It 
provided the scientific community with a method to build realistic 3D human 
body meshes of different shapes in different poses. In order to synthesize a 
human body mesh a template body must be deformed. The model exhibits, however, 
an important limitation, namely the last 
step on the pipeline is an expensive optimization problem. Other researchers 
had used variation of the SCAPE model (S-SCAPE \cite{Pishchulin.2017}) to 
synthesize human bodies but focused on people detection.
After some attempts on improving the human models quality, for example, to make 
more 
easily capturing the correlation between body shape and pose 
\cite{HaslerSSRS09}, or to better understand body dimensions (Semantic 
Parametric Reshaping of Human Body Models - SPRING model 
\cite{Yang.2014}), Loper et al., 2015 \cite{Loper.2015} developed the SMPL 
generative 
human body model. In subsection \ref{subsec:3d_mesh_synth} we synthesize 3D 
human meshes using this 
model. Our approach to synthesize images from 3D meshes has been influenced 
by the recent publication of the large scale dataset SURREAL 
\cite{varol17_surreal}. This work uses the SMPL model 
to generate images from humans with random backgrounds. However, no human body 
dimensions are 
computed or estimated.

More recently, other models have been presented: \cite{Joo.2018} introduced a 
human model with added hands and face and \cite{STAR:ECCV:2020} propose STAR, 
which is trained on a dataset of 14,000 human subjects. We do not use these 
more 
complex models because the human dimensions we estimate do not require such 
level of detail.

\subsection{Digital Anthropometry on 3D Models.}\label{subsec:da}

Although extensive research has been carried out on human shape estimation, 
methods to consistently define shape based on 3D mesh 
anthropometric measurements have been little explored. Only a handful of 
researchers 
have 
reported calculating 1D body dimensions from 3D triangular meshes to use them 
as ground truth for training and validation in a variety of inference processes.

Early research performed feature analysis on what they 
call body intuitive controls, e.g. height, weight, age and gender 
\cite{Allen.2003}  but they do not 
calculate them. Instead they use the CAESAR demographic data. Recording human 
dimensions beyond height like body fat and the more abstract ``muscles" are 
described by \cite{HaslerSSRS09}.

Pertinent to our investigation is also the inverse problem: generating 3D human 
shapes from traditional anthropometric measurements \cite{Wuhrer.2013}. Like 
this work we use a set of 
anthropometric measurements that we call dimensions, unlike them we calculate 
1D measurements from 3D human bodies to use them as ground truth for later 
inference.

Strongly related to our work are methods that calculate waist and 
chest circumference by slicing the mesh on a fixed plane and compute the convex 
hull of the contour \cite{Guan.2013} or path length from identified (marked) 
vertices on the 3D mesh \cite{Boisvert.2013,Dibra.2016a,Dibra.2016b, 
	Piccirilli2018}. However, 
is not clear how they define the human body dimensions. 

In contrast, we do not 
calculate the dimensions from fix vertices on the template 
mesh. Instead, we adopt a more anthropometric approach and use domain knowledge 
to calculate these 
measurements.

\subsection{Human Shape Estimation with Artificial Neural 
	Networks}\label{subsec:hse}

A huge amount of research has been conducted in recent years to address the 
problem of 3D/2D human shape estimation using CNNs. State-of-the-art methods 
are \cite{kanazawaHMR18, wang2020neural, alldieck2019tex2shape} where they 
estimate human 
shape and pose (recently with evolutionary techniques \cite{Li_2020_CVPR}) or 
directly regress 3D vertex coordinates \cite{kolotouros2019cmr}. 
While these CNN's output are human body models parameters 
(i.e., $\beta$s in \cite{Dibra.2016a} and \cite{varol18_bodynet}) or 3D human 
pose (i.e., 3D human key-point coordinates) or both \cite{ExPose:2020} our 
network is capable to output 
human dimensions directly, enabling new perspectives in HSE.

\section{Approach}
\label{gen_inst}
In order to train our neural anthropometer with images of humans and be able to 
use 
human 
body dimensions as ground truth, we base our approach on three main aspects: a) 
3D 
human mesh synthesis with the SMPL model \cite{STAR:ECCV:2020}, b) calculation 
of human body dimensions and c) 2D image synthesis of the generated meshes. 
While a) and c) have been treated and employed by previous work, our 
fundamental contribution lies in b) and the CNN that we train to conduct HBDE. 
We 
now 
turn to discuss these aspects.
\subsection{3D human mesh synthesis with SMPL 
	model}\label{subsec:3d_mesh_synth}
Basically, we use SMPL as generative model because it is derived from real 
humans. This feature guarantees, to a large extent, the 3D mesh level of 
realism and, correspondingly, the correctness of the human body dimensions we 
calculate. Thereof, SMPL has been widely adopted by researchers and 
industry and is the model of preference in HSE.

The SMPL model contains a template mesh (mean template shape 
$\mathbf{\bar{T}}$ in the zero pose $\vec{\theta}^*$) that can be 
deformed according to shape parameters $\vec{\beta}$. Both template and the 3D 
models resulting from synthesis have a mesh resolution of $N = 6890$ vertices. 
Additionally, SMPL attaches a skeleton to the template mesh. The 3D model can 
then be posed according to pose parameters $\vec{\theta}$. This feature plays 
an important role in our methodology since HBDs are commonly defined based on 
skeleton joints. The skeleton $\mathbf{J}$ is defined by its joints location in 
3D 
space $j_i \in \mathbb{R}^3$  and its kinematic tree. Two 
connected 
joints define a ``bone". Moreover, a child bone is rotated relative to 
its 
connected parent. The pose description $\vec{\theta}$ is the specification of 
every bone 
rotation plus an orientation for the root bone.

When synthesizing the human meshes, we aim at gaining high 
variability of human bodies. However, how to choose shape parameters 
$\vec{\beta}$  \emph{such as the variability in the data is as high as 
	possible} is not trivial. If the shape parameters are not 
carefully chosen, either the human meshes look notably similar or so called 
monster shapes might be generated. Understandably, we want to avoid such 
monster shapes, even at the cost of sacrificing variability. Other researchers 
have avoided this problem by fitting the SMPL model to 
already existing 3D meshes \cite{varol17_surreal}. But this implies collecting 
the 3D meshes in the first place and performing expensive optimizations. We 
circumvent this phase by uniformly varying shape parameters from $-3$ to $3$ 
standard deviations. 

Beside shape, we have to consider pose $\vec{\theta}$ as well, as for a number 
of HBDs (in real life) a person is asked to adopt a pose other than the zero 
pose $\vec{\theta}^*$. 
Intuitively, to compute our HBDs, we would like to mimic the procedure 
conducted by a tailor or anthropometrist.
Although these two procedures differ slightly, for the measurements we 
investigate, the subject must lower her arms (shoulders need to hang naturally 
in a 
relaxed 
position). However, we observe that if we impose this in general for all 
3D scans, and due to the fact that we are working with Linear Blend Skinning  
(which produces
artifacts) inter-penetrations occur. Therefore, we bend the arms (set both 
shoulder rotation) by $30$ degrees and elbows by $18.75$ degrees. Following the 
convention that the template mesh is in the zero pose $\vec{\theta^*}$, we 
define this new pose as \textbf{pose 1} 
($\vec{\theta^1}$), see 
Fig. \ref{fig:na_shoulder_width}.

We then synthesize 6000 female and 6000 male 3D meshes in poses {$\vec{\theta^* 
	}, \vec{\theta^1}$} and proceed with the HBDs' calculation.

\subsection{Computation of HBDs}

Regarding our HBDs' calculation, our general approach is to use ray casting 
originating in skeleton joints to establish 
landmarks on the 3D mesh. All of our landmarks arise as a result of a cast ray 
intersecting the mesh in a specific point. In fact, these landmarks allows us 
to perform length computations. For example, the 
length of a subcurve that connects specific landmarks. Nonetheless, length 
queries 
and curve fitting methods on manifolds are in general, and specifically on 3D 
shapes, highly complex. Therefore, we pursue solutions that while 
simulating the way measurements are taken in real life, simultaneously are as 
efficient as possible.

Using landmarks and bounding box associated points, we define 
planes to intersect the mesh. The result of an intersection may be a 
collection of curves both open and closed. Let us restrict our attention to  
closed curves that possibly contain two landmarks. These 
landmarks 
separate the closed curve in two 
subcurves.
We can then make certain assumptions and associate the length of a subcurve 
with the desired HBD. We now define all required landmarks (see Fig. 
\ref{fig:na_arm_length}) to 
explain HBD computation bellow.

\begin{itemize}
	\item $u_{rs}$ and $u_{ls}$ generated by the mesh intersection of a ray from
	right and left \textbf{shoulder 
		joints}, in direction perpendicular to the rear edge of 
	the 
	bounding 
	box's top face, respectively.
	\item $u_{re}$ and $u_{le}$ generated by the mesh intersection of a ray 
	from right and left \textbf{elbow 
		joints}, in direction perpendicular to the right and left 
	bounding 
	box's face, respectively.
	\item $u_{rw}$ and $u_{lw}$ generated by the mesh intersection of a ray 
	from right and left \textbf{wrist 
		joints}, in direction perpendicular to the right and left bounding 
	box's face, respectively.
	\item $u_{ch}$ generated by the mesh intersection of a ray with the 
	\textbf{pelvis joint} in direction 
	perpendicular to the floor (bounding box's bottom face).
\end{itemize}

Additionally, we calculate subject height $h$ (stature), without further 
elaboration, 
as the bounding box's height.

\subsubsection{Shoulder width}
\begin{figure}
	\includegraphics{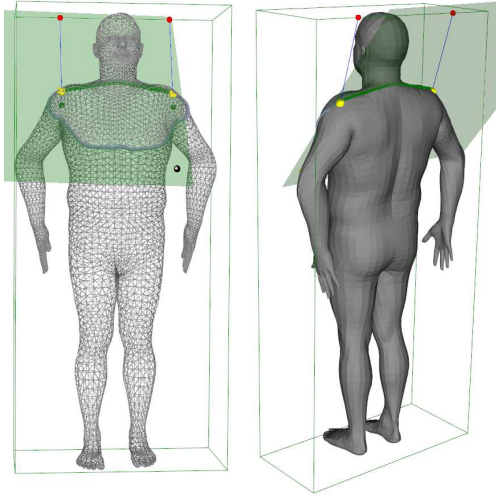}
	\caption{Shoulder width calculation. The 3D (male) mesh is in \textbf{pose 
			one} $\vec{\theta^1}$. Green spheres inside the mesh are skeleton joints. 
		Yellow spheres are 
		skin landmarks (here $u_{rs}$ and $u_{ls}$) determined by casting a ray 
		from the corresponding skeleton joint perpendicular to the bounding box. 
		Red spheres are ray intersection points with the bounding box. The point 
		$p_c$ is the black sphere (slightly translated for visualization purposes). 
		The plane $\pi_{sw}$ (light green) defined by   
		$u_{rs}$, $u_{ls}$ and $p_c$ cuts the mesh and delimits a boundary curve. 
		On the left side we show the subject frontal view, where the boundary curve 
		can be seen. On the right side, landmarks $u_{rs}$ and $u_{ls}$ can be used 
		to further split the curve in two subcurves. We assume that the least 
		length subcurve is the shoulder width (green bold subcurve). The resulting 
		shoulder width for this male subject ($height = 181.11 \, cm$) is $47.60 \, 
		cm$}
	\label{fig:na_shoulder_width}
\end{figure}

For this body dimension, we use landmarks $u_{rs}$ and $u_{ls}$ defined 
above, 
see Fig. \ref{fig:na_shoulder_width}. 
Intuitively, we would extend a virtual tape over the 
mesh surface (skin) to measure the length of a path joining these two points. 
In order to complete the solution, we must further constraint the path to live 
in a subcurve resulting from a plane $\pi_{sw}$ containing the landmarks.

To construct the plane three non-colinear points must be defined; landmarks 
$u_{rs}$ and 
$u_{ls}$ are two of them. However, we need a third point $p_c$. A key insight 
is that plane $\pi_{sw}$ should not be parallel to any of 
the bounding box's faces, as the generated curve would be either too low or two 
high in the back of 
the subject. Therefore, after some experiments we came up with a heuristically 
determined point $p_c$.
The $y$-coordinate is set to be $65$\% of the height, the $x$-coordinate refers 
to the symmetry axis of the body (middle), 
and the $z$-coordinate is given by the maximal $z$-value of the complete mesh 
("thickest point"). As a consequence, we get a plane, which is adapted to the 
body shape.

We slice the mesh with $\pi_{sw}$ and obtain the 
boundary curve. We merely have to decide 
which of the subcurves corresponds to the shoulder width. We do that by 
computing 
the path length 
of the two subcurves and assuming the shoulder width to be the shorter path 
length. 

\subsubsection{Arm length (sleeve length)}
\begin{figure}
	\includegraphics{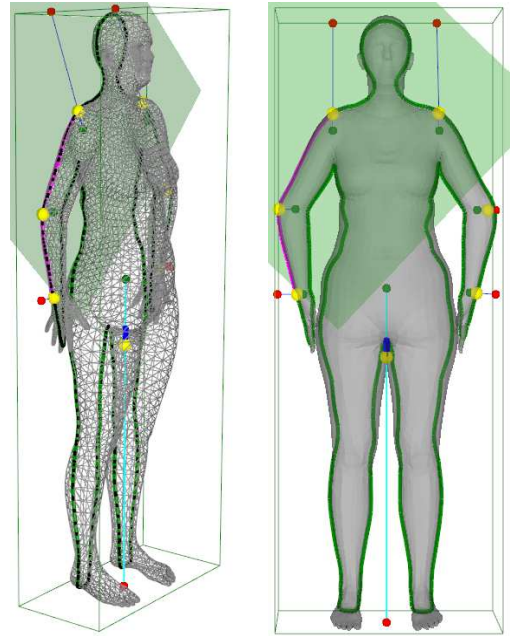}
	\caption{Right arm length (magenta subcurve) and inseam length (cyan line) 
		calculation. Green, yellow and red spheres representation is as described 
		in Fig. \ref{fig:na_shoulder_width}. Following landmarks are shown: from 
		subject's view, starting at the 
		crotch region, clockwise: $u_{ch}$, $u_{lw}$, $u_{le}$, $u_{ls}$, $u_{rs}$, 
		$u_{re}$ and $u_{rw}$. The light green plane 
		defined by three of the corresponding same side landmarks (shoulder, elbow 
		and wrist) cuts the mesh and defines a boundary curve. Left: 
		intersection points of the plane with mesh triangles (edge or face) are 
		shown in black. Right: subject frontal view. Two skin 
		landmarks (here $u_{rw}$ and $u_{rs}$) can be used to further split the 
		curve in two subcurves. We assume that the subcurve being proxy to the 
		corresponding arm length has minimum $x$ coordinate (right), respectively, 
		maximum $x$ coordinate (left). To compute 
		the inseam, we cast a ray in direction to the floor. Note that this ray 
		might intersect the mesh in more than one point 
		(shown in blue). We define the inseam as the length from the last point of 
		this set ($u_{ch}$) to the floor. The resulting right arm and inseam 
		length for this 
		female subject ($height = 196.05 \, cm$) are $64.67 \, cm$ and $85.72 \, 
		cm$, 
		respectively.}
	\label{fig:na_arm_length}
\end{figure}

Similarly to how we calculate shoulder width, we proceed with arm length 
calculation. In this subsection, we describe the method for right arm length. 
The left arm length calculation is handle using exactly the same 
methodology, albeit with corresponding landmarks (see Fig. 
\ref{fig:na_arm_length}).

For this dimension we use landmarks $u_{rs}$, $u_{re}$, and $u_{rw}$. A plane 
$\pi_{ral}$ defined by these three points cuts the mesh and yields a number of 
polygonal curves (possibly open and closed). We observe that this step produces 
frequently 
more than a single curve. Therefore, we are forced to explicitly search for the 
right curve containing the landmarks.

Theoretically, this search should not pose further complications. In practice, 
floating point operations are an important issue due to their limited 
precision. Thus, we introduce a tolerance parameter $tol = 0.001$ and deem a 
point to be found when it is within the tolerance.

Once we find the desired curve, we proceed in the same manner as for the 
shoulder width calculation: we use $u_{rs}$ and $u_{rw}$ to extract two 
subcurves. We then can consider the right arm length to be the subcurve with 
minimum $x$ coordinate and calculate its length.

Not surprisingly, this method does not yield equal right and left arm length. 
Curiously, despite our computations being conducted on 3D 
meshes, this type of asymmetry is also observed in modern 
humans\cite{AUERBACH2006203}.

\subsubsection{Inseam}
The inseam is informally defined as the length of the shortest path 
between the crotch (perineum) and the floor. Therefore, we calculate the 
euclidean distance from landmark $u_{ch}$ to the bounding box's bottom face, 
see Fig. \ref{fig:na_arm_length}.

At this point, we have calculated our four HBDs. Since Calvis use the SMPL 
model as well, we are able to easily calculate the other three HBDs, therefore, 
completing our formulation.

\subsection{2D image synthesis for network input}
After having calculated the HBDs, we continue with the synthesis of the neural 
anthropometer's input: 2D images of the 3D meshes.

For the synthesis we use the Blender package \cite{blender}, adapting the 
method 
described in 
previous work \cite{varol17_surreal, ygtham2020calvis}. Specifically, we employ 
Cycles rendering 
engine with an orthographic camera model. The orthographic camera has a 
resolution of $200 \times 200$, focal 
length of $60 \, mm$, sensor size of $32 \, mm$ and orthographic scale of 
$2.5$. We keep fix the camera at a distance of $6 \, m$ in front of the mesh 
and produced a grayscale image.

Having synthesized input and supervision signal, we are now ready to detail the 
neural anthropometer architecture.

\subsection{Neural Anthropometer Architecture}\label{subsec:na_arqui}

The neural anthropometer architecture is shown in Fig. \ref{fig:na_overview} 
and we implement it with 
Pytorch \cite{pytorch_2019}. Our intention is to keep the 
network as small as possible, since  
such a network is easier to train and consumes less resources . The input 
layer processes a fixed image of size $\mathtt{200 \, x \, 200 \, x \,1}$ by 
applying a convolution with a 5-pixels square kernel to produce a feature map 
of size $\mathtt{196 \, x \, 196 \, x \,8}$. The intuition behind choosing the 
number of channels is that the network might need at least one channel per HBD 
in the first stage to achieve good performance. The tensor is then passed 
through a rectified linear unit (ReLU) \cite{relu} and batch normalization is 
applied. 
Next, max pooling with stride $2$ is used before the tensor is send 
to a second 
convolutional layer with a $5$-pixels square kernel and $16$ output channels, 
resulting in a tensor 
of size $\mathtt{94 \, x \, 94 \, x \, 16}$. The same pooling strategy as 
before is applied and the output is flatten to a tensor of size 
$\mathtt{35344}$. This tensor is passed to a fully connected layer and again 
through a ReLU. The last layer is a regressor that outputs the eight human body 
dimensions in meters.

\section{Experiment and Results}\label{sec:exp_and_results}

\begin{figure*}
	\begin{center}
		\includegraphics{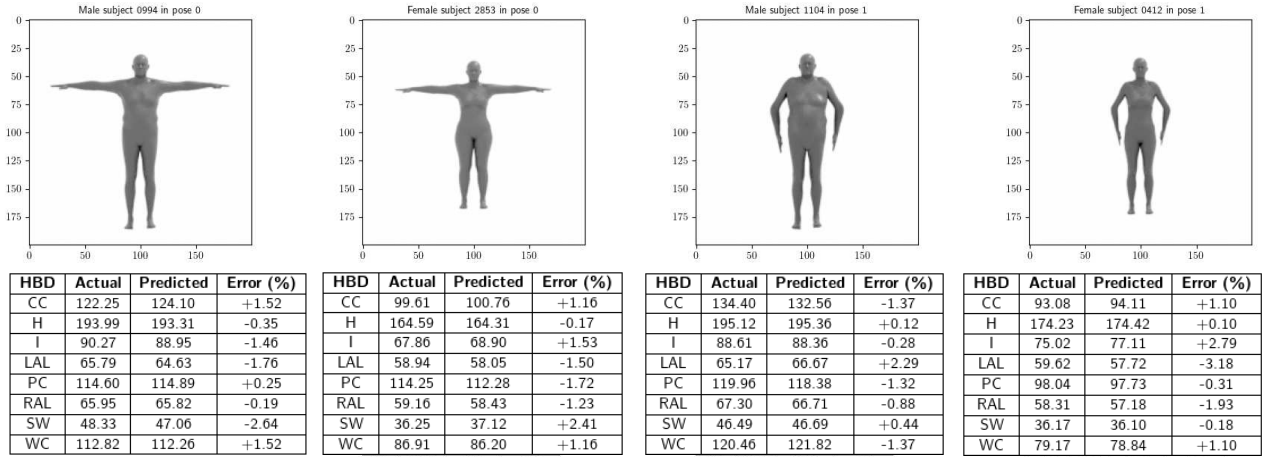}
	\end{center}
	\caption{Estimation results on four selected subjects, two female and two 
		male. Abbreviations 
		are as follows. CC: chest circumference, H: height, I: inseam, LAL: left 
		arm length, PC: pelvis circumference, RAL: right arm length, SW: shoulder 
		width, WC: waist circumference.}
	\label{fig:est_results}
\end{figure*}
We suppose that the neural anthropometer can infer the HBDs from images of 
humans in 
different poses.
To better understand this phenomenon, we train and evaluate the 
network with all images, regardless of gender or pose, in a k-fold 
cross-validation setting ($k = 5$).

We conduct our experiment with the neural anthropometer on the dataset we 
produced. Let us briefly recapitulate its structure. The dataset 
comprises $12000$ human body meshes from $6000$ subjects 
generated with SMPL: $3000$ female meshes in \textbf{pose 0} and $3000$ in 
\textbf{pose 1}; equivalently,  $3000$ male meshes in \textbf{pose 0} and 
$3000$ in \textbf{pose 1}. From each mesh, we synthesize a grayscale image of 
size $\mathtt{200 \, x 
	\, 200 \, x \,1}$. Therefore, the dataset contains $12000$ synthetic 
pictures 
of the subjects. Further, for each subject, we calculate a set of eight HBDs: 
shoulder width, right and left arm length, inseam; chest, waist and pelvis 
circumference and height.

For training, we randomly selected images of female and male objects in 
\textbf{pose 0} and \textbf{pose 1}. We train the neural anthropometer for $20$ 
epochs and 
use mini-batches of size $100$. We minimize \textbf{MSE} between the 
actual 
and the estimated HBDs using stockading 
gradient descent with momentum (Pytorch's method modified from 
\cite{sgd_with_momentum}). We set learning rate to $0.01$ and momentum to $0.9$.
We perform this experiment on a PC with Intel(R) Core(TM) i7-7700K CPU and 
NVIDIA GeForce GTX 1060Ti.

\subsection{Quantitative Evaluation}
As model evaluation we use $k\text{-}$fold cross validation ($k = 
5$), e.g., each fold $j$ contains $a = 2400$ instances for evaluation, from 
which 
we 
estimate eight HBDs using the 
neural anthropometer. Therefore, we report statistics based on a results tensor 
of shape
\begin{equation}
	k \, \times \, a \, \times \, |\{\hat{D_i}, D_i\}| \, \times \, 8 = 
	\mathtt{5 \, x \, 2400 \, x \, 2 \, x \, 8 }.
\end{equation}

We report estimation error $e^{i}_{MAD}$ for 
each HBD 
$i$ as the 
Mean Absolute Difference over the $j$ folds between 
estimated and actual HBDs $\hat{D_i}$, $D_i$, respectively:

\begin{align}
	e^{j}_{MAD} & = \frac{1}{a}\sum_{l=1}^{a} \lvert 
	\hat{D_l} - D_l \rvert, &
	e^{i}_{MAD} & = \frac{1}{k}\sum_{j=1}^{k} e^{j}_{MAD}
\end{align}

Additionally, we find helpful to consider Relative Percentage Error (RPE)
$e^{i}_{RPE}$ for each HBD $i$ and its Average (ARPE);

\begin{align}
	e^{j}_{RPE} & = \frac{1}{a}\sum_{l=1}^{a} \lvert 
	\frac{\hat{D_l} - D_l}{D_l} \rvert, &
	e^{i}_{RPE} & = \frac{1}{k}\sum_{j=1}^{k} e^{j}_{RPE}
\end{align}

Like prior research, we report Average Mean Absolute 
Difference (AMAD) over the eight HBDs $e_{AMAD} = \frac{1}{8}\sum_{j=1}^{8} 
e^{i}_{MAD}$.

\begin{table}
	\begin{tabular}{@{}lll@{}}
		\toprule
		HBD & MAD & RPE(\%) \\
		\midrule
		Shoulder width  & 12.54 & 4.93 \\
		Right arm length & 12.98 & 2.22 \\
		Left arm length & 13.48 & 2.34\\ 
		Inseam/crotch height & 22.17 & 3.12\\ 
		Chest circumference & 25.22 & 2.51\\ 
		Waist circumference & 27.53 & 3.67\\ 
		Pelvis circumference & 25.85 & 2.40\\ 
		Height & 27.34 & 1.58 \\
		\midrule
		AMAD & 20.89 & \\
		ARPE &  & 2.84 \\
		\bottomrule
	\end{tabular}
	\caption{Neural Anthropometer's estimation results. Displayed are the MAD 
		for each human body dimension (first column) over five folds, expressed in 
		millimeters (second column). The third column shows the RPE. The last two 
		rows display AMAD and ARPE over all HBDs and folds.}
	\label{table_results}
\end{table}

Results are shown in Table \ref{table_results}. From the table, it can be 
seen 
that the prediction of shoulder width exhibits the highest relative 
estimation error ($4.93$\%), while height prediction manifests lowest relative 
error ($1.58$\%). Interestingly, right and left arm length prediction error 
($2.22$\% and 
$2.34$\%, respectively) are closest to 
each other than to any other estimation error. Overall, the neural 
anthropometer is able to predict the HBDs with average 
relative error of 2.84\%.

\subsection{Visualization and Discussion}

Let us visualize a four instances mini-batch. We select randomly one of $k$ 
models saved during training to make inference. Fig \ref{fig:est_results} shows 
human body dimensions estimation of four selected subjects.

Fig \ref{fig:est_results} is quite revealing in several ways. First, consider 
the first two subjects (male subject 0994 and female subject 2853). Both 
subjects are in pose 0 and their shoulder width prediction error ($-2.64$\% and 
$2.41$\%) is higher than 
the other two (in pose 1). This may indicate that for network is not easy to 
predict shoulder width when the person is in pose 0. Secondly, as it might be 
expected, height exhibits across all subjects and HBDs the lowest error 
($-0.35$\%, 
$-0.17$\%, $+0.12$\% and $+0.10$\%). This result has further 
strengthened our confidence in considering the height as the less difficult 
predictable HBD. Finally, while the right and left arm length is mostly 
underestimated, the error is slightly higher in the case of left arm length.

We believe that it might not be reasonable to improve some of the results we 
present 
here. For example, the difference between estimated and actual height 
is in the order of mm. That means that estimation is already being performed 
at a precise level. Probably, in most scenarios, this level of sensitivity is 
not needed. On the 
other hand, it is practical to improve estimation error for other 
HBDs like arm length. Here, we observe that the estimation error is probably 
relevant for applications like distance 
tailoring. This suggests that, depending on the context, AMAD might not be 
optimal to evaluate results, 
since the estimation error of the HBDs are equally weighted. A weighted average 
could be more desirable when reporting results.

\section{Conclusion}
We presented the neural anthropometer, a CNN capable of estimating eight human 
body dimensions from synthetic pictures of humans. 
To obtain ground truth for training, we introduced a method to calculate four 
human body 
measurements (shoulder width, right and left arm length, and height) from 3D 
human meshes. We then augmented this set with four human body 
dimensions calculated using recently published methods. This method constitute 
an innovation that can be used by 
researchers to push the boundaries of Human Shape Estimation. We then input the 
images to the CNN and minimized the Mean Square Error between the estimated and 
the calculated human body dimensions.
Our experiments revealed that the neural anthropometer is able to accurate 
estimate eight human body measurements with a mean 
absolute difference of $20.89$ mm (relative error of 2.84\%).

\bibliography{egbib}
\bibliographystyle{ieeetran}

\end{document}